\documentclass[10pt,twocolumn,letterpaper]{article}
\usepackage{amsmath}
\usepackage{algorithm}
\usepackage{algorithmic}
\usepackage{booktabs}
\usepackage{iccv}
\usepackage{times}
\usepackage{epsfig}
\usepackage{graphicx}
\usepackage{amsmath}
\usepackage{amssymb}
\usepackage{subfigure}
\usepackage{amsfonts}

\usepackage[breaklinks=true,bookmarks=false]{hyperref}

\iccvfinalcopy 

\def\iccvPaperID{****} 

\ificcvfinal\fi

\begin{document}

\title{EM-Net: Gaze Estimation with Expectation Maximization Algorithm}



\author{
Zhang Cheng$^1$, Yanxia Wang$^1$, Guoyu Xia$^1$\\
$^1$ School of Computer and Information Science, Chongqing Normal University, Chongqing, 401331, China\\
{\tt\small 2022210516040@stu.cqnu.edu.cn(Z. Cheng), wangyanxia@cqnu.edu.cn(Y. Wang)} \\ 
{\tt\small 2023210516082@stu.cqnu.edu.cn(G. Xia)}
}

\maketitle

\begin{abstract}
In recent years, the accuracy of gaze estimation techniques has gradually improved, but existing methods often rely on large datasets or large models to improve performance, which leads to high demands on computational resources. In terms of this issue, this paper proposes a lightweight gaze estimation model EM-Net based on deep learning and traditional machine learning algorithms \textbf{E}xpectation \textbf{M}aximization algorithm.  First, the proposed \textbf{G}lobal \textbf{A}ttention \textbf{M}echanism (GAM) is added to extract features related to gaze estimation to improve the model's ability to capture global dependencies and thus improve its performance. Second, by learning hierarchical feature representations through the EM module, the model has strong generalization ability, which reduces the need for sample size. Experiments have confirmed that, on the premise of using only 50\% of the training data, EM-Net improves the performance of Gaze360, MPIIFaceGaze, and RT-Gene datasets by 2.2\%, 2.02\%, and 2.03\%, respectively, compared with GazeNAS-ETH. It also shows good robustness in the face of Gaussian noise interference.
\end{abstract}

\section{Introduction}
Gaze is a form of non-verbal communication that can convey emotions and express intentions.  Gaze estimation refers to inferring the direction of eye gaze through relevant techniques. With the gradual development and advancement of gaze estimation techniques, it has been widely used in the fields of medical diagnosis\cite{1}, online examination\cite{2}, driver attention analysis\cite{3}, and human-computer interaction \cite{4,5}. For different application scenarios, high-precision and lightweight gaze estimation models are still of great research significance.

Gaze estimation methods can be categorized into two types: model-based and appearance-based. Model-based gaze estimation methods \cite{6,7,8,9} use an eye model to estimate gaze direction, which fits geometric features such as pupil center, eye position, iris contour, etc. However, this type of method requires an accurate model of the eye and suffers from high computational complexity and poor adaptation. Appearance-based gaze estimation methods use full-face or eye images to infer gaze direction. Appearance-based methods can be categorized into two types based on the input data: eye image-based \cite{10,11,12}  and full-face image-based \cite{13,14,15,16}. Eye image-based methods use either monocular or binocular images as input data, and since the direction of gaze is closely related to the head pose, these methods usually need to concatenate the head pose information after extracting the eye features to compute the final gaze direction. The full-face image-based methods directly use face images as input, which contains more features related to gaze estimation and can effectively improve the performance of the model. Therefore, this method has gradually become a research hotspot.

With the development of CNN, the performance of gaze estimation methods based on full-face images has steadily improved, but the models are also getting larger and larger, so larger datasets or appropriate regularization methods are needed to match them. In addition, large models usually have problems such as high complexity, high power consumption, and difficult to train and deploy. On the other hand, existing gaze estimation methods have a small receptive field that cannot fully utilize the global information to achieve long-distance modeling and do not have a good solution for situations such as incomplete or occluded data. Therefore, the application of methods that utilize large models to improve accuracy is limited.

Various lightweight networks have been proposed to alleviate this problem to some extent, such as MobileNetV3 \cite{17} and Mobilevitv3 \cite{18}. Xu et al. \cite{19} use Mobilevitv3 as the backbone network and combine it with a fast Fourier transform to propose a lightweight model FR-Net, which achieves high accuracy. However, due to the large resolution of the input image, the computational cost of the model increases and the inference time becomes slow. In terms of the above issue, this paper proposes a lightweight gaze estimation model EM-Net based on full-face images. The model takes the improved MobileNetV3 as the backbone network and combines with a machine learning algorithm the Expectation Maximization(EM) algorithm to solve the problem of incomplete data or occlusion.

The main contributions of this paper are as follows:
\begin{itemize} 
\item A new Global Attention Mechanism (GAM) is proposed, which can not only effectively integrate spatial information and channel information but also significantly expand the receptive field of the model and capture long-distance dependencies, to make more comprehensive use of the features in the sequence and improve the overall performance of the model.
\item Based on the EM algorithm, an EM module is proposed, which retains the characteristic that the EM algorithm is good at processing unobserved variables, and can further refine the feature representation based on the features extracted by CNN to better reflect the statistical characteristics and potential structure of the data, to make the processing of new data more robust and accurate.
\item A new lightweight gaze estimation model EM-Net is proposed, which not only significantly reduces Parameters and FLOPs, but also achieves excellent performance using only 50\% of the training data from the public datasets Gaze360, MPIIFaceGaze, and RT-Gene. 
\end{itemize}

\begin{figure*}
	\centering
\includegraphics[width=.9\textwidth]{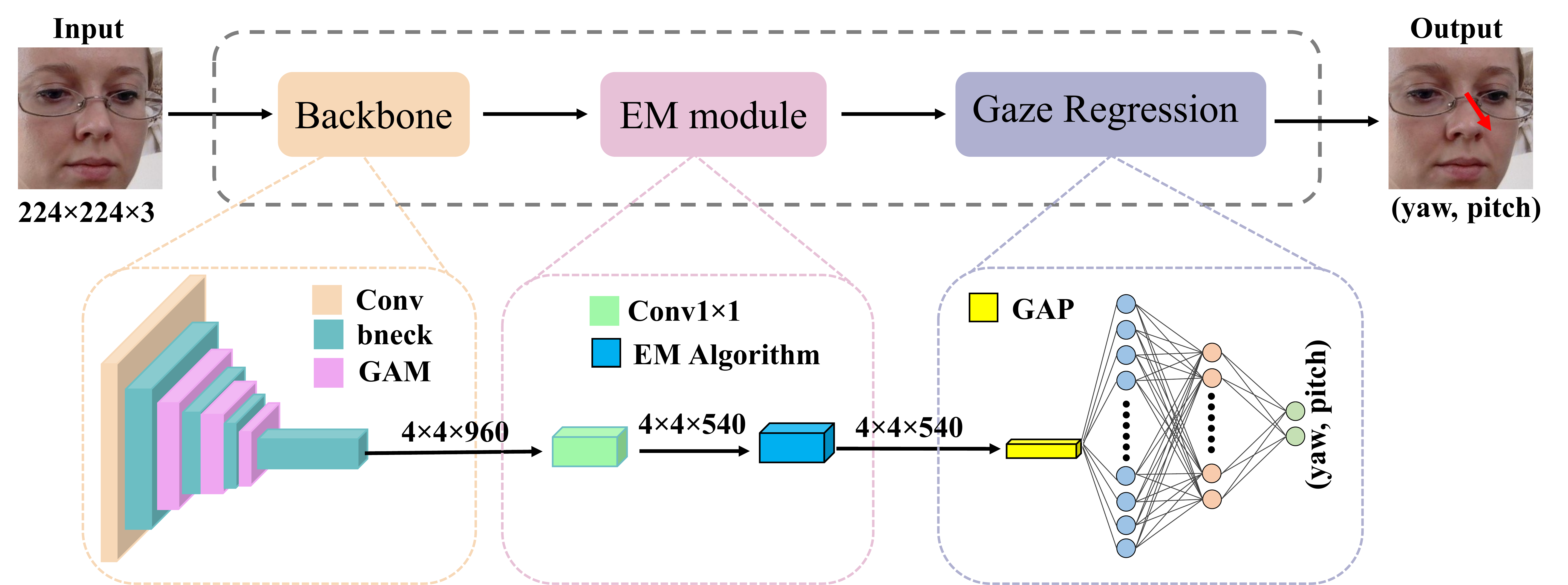}
	\caption{EM-Net.}
	\label{EM_Net}
\end{figure*}

\section{Related work}
This section mainly introduces the gaze estimation task using a full-face image as input, attention mechanism, and EM algorithm.
\subsection{Full-Face Appearance-Based Gaze Estimation}
Zhang et al.  \cite{20} propose a full-face image gaze estimation method based on spatial weight mechanism. The main idea is to use the spatial weight mechanism to encode the weights of different regions of the whole face, suppress the regions that do not contribute to the gaze estimation task, such as background, and enhance the weight of the eye region. Cheng et al.  \cite{13} believe that among the rich information in facial images, only the features of the eye area are related to gaze, while other features are unrelated to gaze, such as facial expressions, lighting, and personal appearance. Therefore, they propose a feature purification based on a self-adversarial framework for gaze estimation, which preserves gaze-related features and removes irrelevant facial image information. However, Abdelrahman et al.  \cite{21} believe that the appearance of the eyes is unique, as well as the diversity of lighting and gaze directions. Therefore, they propose to perform gaze regression on multiple gaze angles separately to improve the prediction accuracy of each angle and overall performance. Cai et al.  \cite{22} believe that model performance degradation can occur due to data and model uncertainty. They propose the Uncertainty Reduction Gaze Adaptation(UnReGA) framework, which reduces sample uncertainty by enhancing image quality, minimizes prediction variance for the same input, and reduces model uncertainty. The performance of the gaze estimation method using facial images as input has been improved, but the model structure is relatively complex and the receptive field is small.
\subsection{Attention Mechanism}
Attention mechanism is a method that mimics the human visual and cognitive system. The introduction of attention mechanism neural network can automatically learn and selectively focus on the important information in the input, improving the performance and generalization ability of the model. According to application scenarios, common attention mechanisms can be mainly divided into multi-head attention mechanisms applied to Transformer architecture \cite{23}, SE \cite{24}, and GCT \cite{25} applied to CNN architecture. The multi-head self-attention mechanism proposed by Vaswani et al. \cite{23} uses multiple independent attention heads to separately calculate the relative importance between each element in a sequence and adaptively captures the long-distance dependence between elements, enabling the model to better understand contextual information. Then the results are concatenated and projected again to get a richer feature representation, to enhance the expressiveness of the model. The SE attention mechanism proposed by Hu et al. \cite{24} integrates spatial information through the squeeze operation, and the channel information interaction is achieved through the Excitation operation. The SE attention mechanism explicitly models the interdependencies between channels, which helps the model to capture richer feature information and thus improve the model performance. The existing attention mechanism can improve the performance of the model, but there are still some shortcomings, such as: only one of channel information and spatial information is used, or both channel information and spatial information are involved, but one of the information is not fully used.
\subsection{Expectation Maximization Algorithm}
The EM algorithm is a commonly used iterative optimization algorithm in machine learning, suitable for situations where data is incomplete or contains hidden variables. By iteratively executing the Expectation step and Maximization steps, the parameters of the probability model containing hidden variables are estimated, and each iteration ensures that the value of the objective function is not reduced, so it usually has good convergence. Hinton et al. \cite{26} combined the EM algorithm with capsule networks to achieve significant results in multiple publicly available datasets, enhancing resistance to white-box adversarial attacks. Zhang et al. \cite{27} propose a lightweight Multi Interest Retrieval Network (MIRN) by introducing EM routing, which adaptively learns vector-guided model training of user representations and achieves significant performance in recommendation systems. In the task of gaze estimation, there is a problem of incomplete facial image information caused by factors such as facial occlusion and lighting conditions. However, the EM algorithm is well-suited for situations with incomplete data or hidden variables. Therefore, this paper proposes to introduce the EM algorithm to the task of gaze estimation.

\begin{table*}[!htbp]
\caption{Comparison of MobileNetV3 before and after improvement}\label{tbl1}
\resizebox{1.0\linewidth}{!}{
\begin{tabular}{|cc|}
\hline
\multicolumn{2}{|c|}{Conv3×3,S=2,P=1,C=16, F,HS} \\ \hline
\multicolumn{2}{|c|}{bneck,3×3,exp\_size=16,C=16,F,RE,S=1}   \\ \hline
\multicolumn{2}{|c|}{bneck,3×3,exp\_size=64,C=24,F,RE,S=2}   \\ \hline
\multicolumn{2}{|c|}{bneck,3×3,exp\_size=72,C=24,F,RE,S=1}   \\ \hline
\multicolumn{1}{|c|}{\textcolor{blue}{\textbf{/}}}  & {\textcolor{blue}{\textbf{GAM}}}                        \\ \hline
\multicolumn{2}{|c|}{bneck,5×5,exp\_size=72,C=40,T,RE,S=2}     \\ \hline
\multicolumn{2}{|c|}{[bneck,5×5,exp\_size=120,C=40,T,RE,S=1]×2} \\ \hline
\multicolumn{1}{|c|}{\textcolor{blue}{\textbf{/}}}   & {\textcolor{blue}{\textbf{GAM}}}   \\ \hline
\multicolumn{2}{|c|}{bneck,3×3,exp\_size=240,C=80,F,HS,S=2}  \\ \hline
\multicolumn{2}{|c|}{bneck,3×3,exp\_size=200,C=80,F,HS,S=1}   \\ \hline
\multicolumn{1}{|c|}{\textcolor{blue}{\textbf{bneck,3×3,exp\_size=184,C=80,F,HS,S=1]×2}}}    &   \textcolor{blue}{\textbf{bneck,3×3,exp\_size=184,C=80,F,HS,S=1}}           \\ \hline
\multicolumn{1}{|c|}{\textcolor{blue}{\textbf{/}}}    &   {\textcolor{blue}{\textbf{GAM}}}       \\ \hline
\multicolumn{2}{|c|}{bneck,3×3,exp\_size=480,C=112,T,HS,S=1}  \\ \hline
\multicolumn{2}{|c|}{bneck,3×3,exp\_size=672,C=112,T,HS,S=1}   \\ \hline
\multicolumn{2}{|c|}{bneck,5×5,exp\_size=672,C=160,T,HS,S=2}     \\ \hline
\multicolumn{1}{|c|}{\textcolor{blue}{\textbf{[bneck,5×5,exp\_size=960,C=160,T,HS,S=1]×2}}}    &        \textcolor{blue}{\textbf{bneck,5×5,exp\_size=960,C=160,T,HS,S=1}}           \\ \hline
\multicolumn{1}{|c|}{\textcolor{blue}{\textbf{Conv 1×1,S=1, C=960, HS}}}   &   \textcolor{blue}{\textbf{/}}      \\ \hline
\multicolumn{1}{|c|}{\textcolor{blue}{\textbf{MLP(960---1280---class\_number)}}}   &  \textcolor{blue}{\textbf{/}}      \\ \hline
\end{tabular}
}
\end{table*}

\section{Methods}
This section elaborates on the EM-Net network model proposed in this paper, which has a large receptive field and is lightweight. It mainly consists of three parts: backbone network, EM module, and gaze regression. The structure is shown in \textcolor{blue}{Fig.\ref{EM_Net}}.

\subsection{Backbone}
\begin{figure}[htbp]
	\centering
	\includegraphics[width=.5\textwidth]{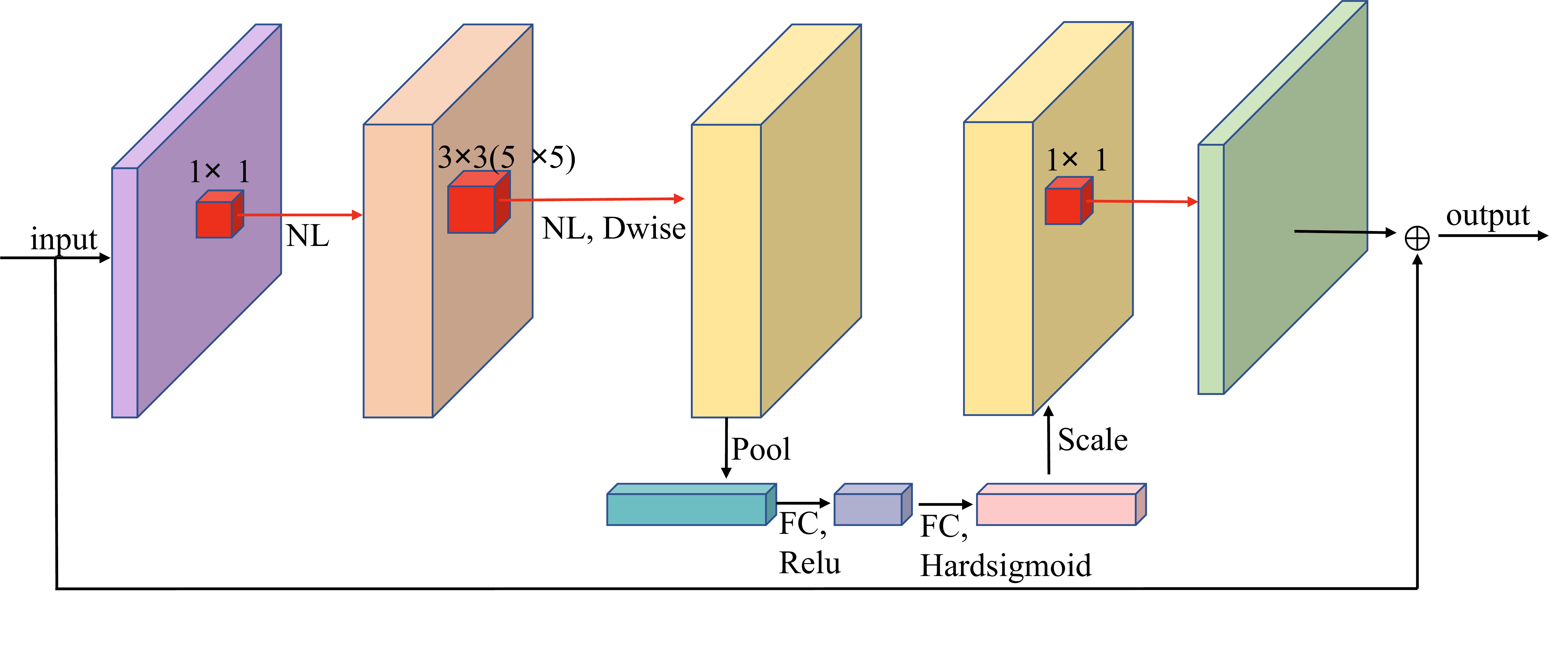}
	\caption{Bneck network structure. NL indicates the activation function, which is different for different layers. For details, see Table 1. Dwise indicates the grouping convolution. When the input channel is equal to exp\_size, there is no 1×1 convolution at the input end to raise the dimension.}
	\label{bneck}
\end{figure}

The backbone network of the EM-Net is improved based on MobileNetV3 \cite{17}, which has fewer layers than MobileNetV3 network and incorporates the \textbf{G}lobal \textbf{A}ttention \textbf{M}echanism (GAM) proposed in this paper. The specific differences between MobileNetV3 and the improved MobileNetV3 are shown in 
\textcolor{blue}{Table \ref{tbl1}}. The left side represents the original MobileNetV3, and the right side represents the improved MobileNetV3. The blue font indicates the areas where the differences exist. In the table, bneck represents the basic structure of the network, as shown in
\textcolor{blue}{Fig. \ref{bneck}}. In the network layers containing the bneck structure, 3x3 or 5x5 represents the size of the convolution kernels for grouped convolutions in bneck, S represents the stride, P represents padding, C represents the number of output channels, F and T represent whether SE attention is used, HS represents the h-swish activation function, RE represents the ReLU activation function, exp\_size represents the number of channels output by the first dimensionally increased 1x1 convolution in bneck, and GAM represents the global attention mechanism.

\begin{figure*}[htbp]
	\centering
	\includegraphics[width=.8\textwidth]{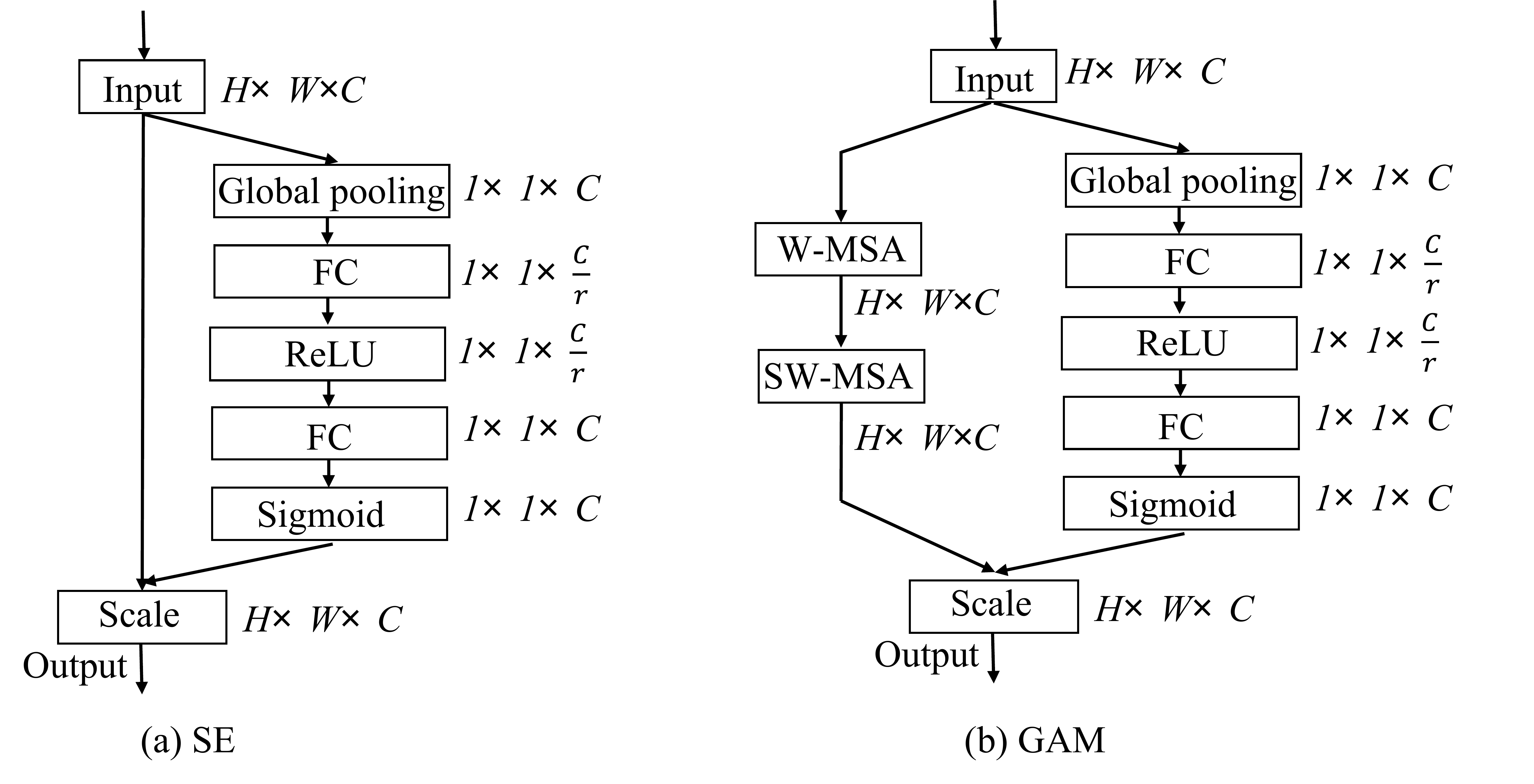}
	\caption{SE and GAM.}
	\label{GAM}
\end{figure*}

\subsubsection{h-swish activation function}
Swish is a nonlinear activation function proposed by Google, which performs better than ReLU in deep network, and has the characteristics of no upper bound and lower bound, smoothness, non-monotonic, etc. Smoothness plays an important role in optimization and generalization, and its calculation formula is as follows:
\begin{equation}
Swish(x)=x*Sigmoid(x)
\end{equation}

h-swish \cite{17} is an improvement based on Swish. Since sigmoid cannot be implemented on hardware devices, sigmoid is replaced with a similar Relu6-based calculation. h-swish is faster than swish and easier to be quantified. The calculation formula is as follows:

\begin{equation}
h-swish[x] = x\frac{ReLU6(x+3)}{6}
\end{equation}

\begin{figure*}[htbp]
	\centering
	\includegraphics[width=.9\textwidth]{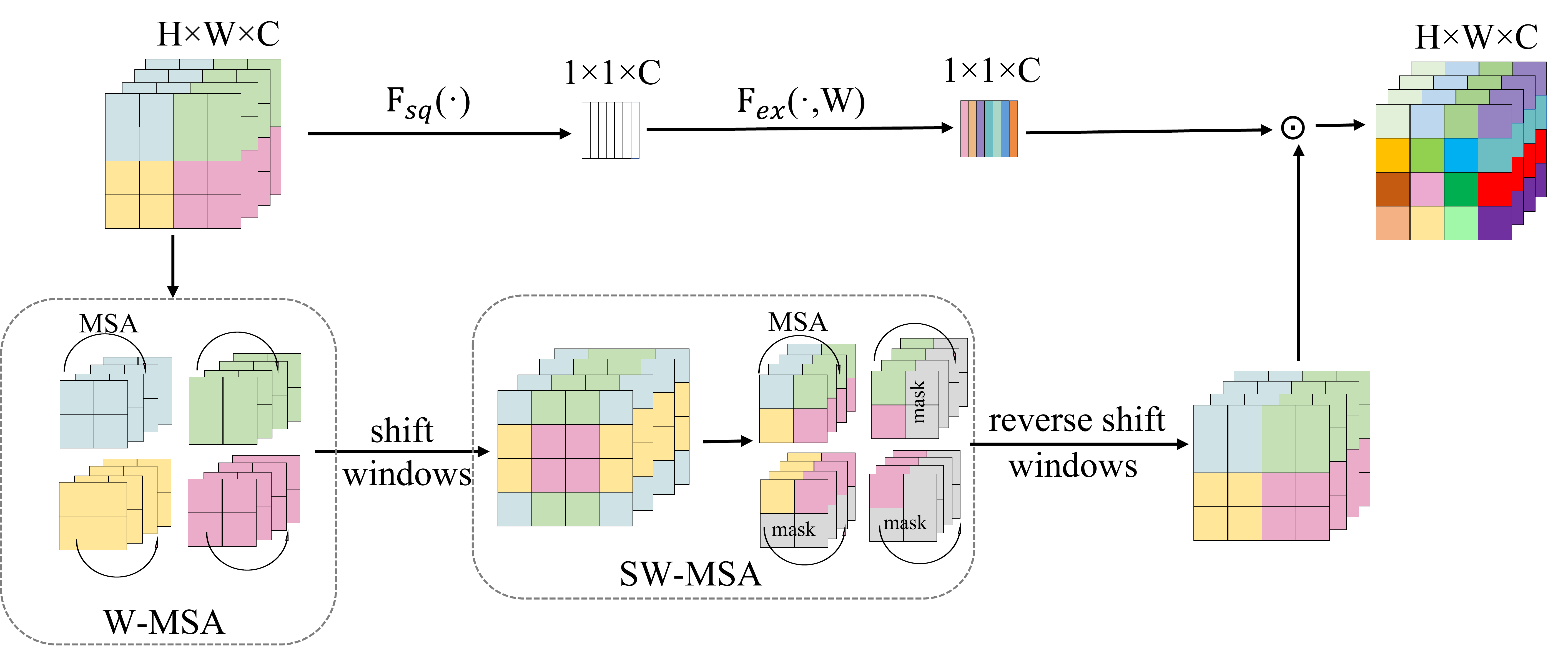}
	\caption{Schematic diagram of GAM information exchange.}
	\label{GAM1}
\end{figure*}

\subsubsection{Global attention mechanism(GAM)}
SE \cite{24} attention can make the model focus more on useful channel information by learning adaptive channel weights. However, SE attention only fuses spatial information through global average pooling, which makes it unable to capture attention in the spatial dimension and does not fully utilize contextual information. In response to these issues, this paper proposes a Global Attention Mechanism (GAM), which not only fully focuses on the connections between channels, but also can capture long-distance dependencies by expanding the receptive field of the model, thereby more comprehensively utilizing contextual information and improving the overall performance of the model. The structure of SE and GAM is shown in \textcolor{blue}{Fig \ref{GAM}}. GAM has two branches. The right branch is responsible for information exchange in the channel dimension, while the left branch is responsible for information exchange in the spatial dimension. The left branch first divides the feature map into multiple windows and performs multi-head self-attention \cite{28} within each window to complete local information exchange. Then, the feature map is moved and the windows are redivided. After adding a mask to the original non-adjacent information in each window, multi-head self-attention is performed again to achieve global information exchange. Finally, the feature map is shifted in reverse and the channel information and spatial information exchange are achieved through matrix dot multiplication. The information exchange of GAM is shown in \textcolor{blue}{Fig \ref{GAM1}}.

\subsection{EM Module}
\begin{figure*}[htbp]
	\centering
	\includegraphics[width=.9\textwidth]{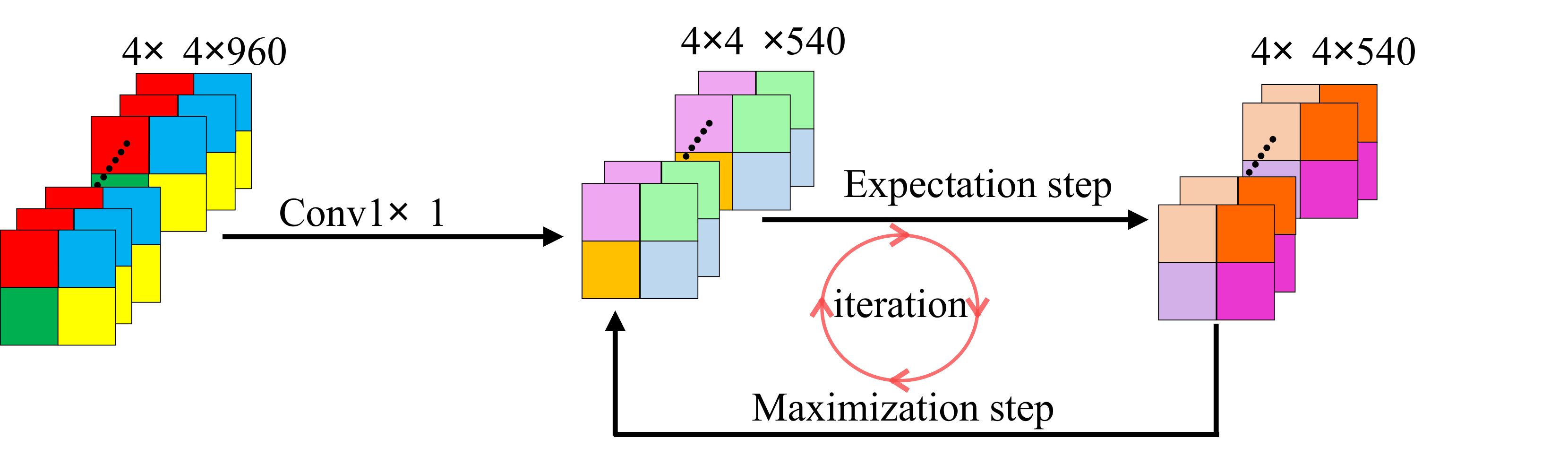}
	\caption{EM Module.}
	\label{EM_Module}
\end{figure*}

The EM module consists of a 1x1 convolutional layer and the EM algorithm. The main function of 1x1 convolution is to reduce the dimensionality of the feature map without significantly increasing the model parameters. After backbone processing, the number of data channels can reach up to 960, while the EM algorithm requires repeated iterations to converge the data, which is time-consuming. If the number of channels is too large, it will cause severe latency and violate the original intention of lightweight networks. Convolutional operations can learn high-level abstract feature representations from raw data, while EM algorithms can leverage the rich information provided by high-level features in parameter estimation to improve the accuracy and robustness of parameter estimation. In addition, EM algorithms help identify information about the potential distribution and intrinsic structure of data, thereby better understanding and processing complex data patterns. The procedure for the computation of the EM module is shown \textcolor{blue}{Algorithm \ref{alg:em}}, and the structure is shown in \textcolor{blue}{Fig \ref{EM_Module}}.

\begin{algorithm}
\caption{EM Module. When the EM module is given the model parameter $\theta^{(i)}$ for the i-th iteration, it returns the model parameter $\theta^{(i+1)}$ for the i+1st iteration.}
\label{alg:em}
\begin{algorithmic}
\STATE \textbf{Input:} Observing variables: $Y$, hidden 
   variable: $Z$, \newline  joint distribution: $P(Y,Z|\theta)$, \newline
  Conditional distribution: $P(Z|Y,\theta)$
\STATE \textbf{Ouput:} Model parameter $\theta$
\STATE  $\theta^{(i)} = \text{Conv1} \times 1$
\FOR{$i$ iteration}
\STATE \begin{equation}
\begin{aligned}
Q\left(\theta,\theta^{\left(i+1\right)}\right) &:= E_Z[logP(Y,Z|\theta)|(Y,\theta^{(i)})] \\ 
     &= \sum_{Z}{logP(Y,Z|\theta)P(Z|Y,\theta^{(i)})} \\
\theta^{(i+1)} &:=arg\max_{\theta} Q(\theta,\theta^{(i+1)}) \\ 
     &= arg\max_{\theta}\sum_{Z}{logP(Y,Z|\theta)P(Z|Y,\theta^{(i)})}
\nonumber   
\end{aligned}  
\end{equation}
\ENDFOR
\RETURN $\theta^{(i+1)}$
\end{algorithmic}
\end{algorithm}

\subsection{Gaze Regression}
The function of Gaze Regression is to infer the gaze direction using the extracted features $f\in\mathbb{R}^{4\times4\times540}$ feature maps are obtained by the EM module, and the $f_1\in\mathbb{R}^{1\times1\times540}$ feature maps are obtained by global average pooling. Then, two fully connected layers are used to regress the 3D gaze direction, which is represented by the horizontal yaw angle and the vertical pitch angle. The calculation formula is as follows:
\begin{equation}
(yaw, pitch)= MLP(GAP(f))
\end{equation}
Where $f$ represents the features input into the Gaze Regression, GAP represents the global average pooling, and MLP represents the multi-layer perceptron used to regress out the three-dimensional gaze direction.

After inferring the yaw angle and pitch angle in EM-Net, the two-dimensional coordinates can be converted into a three-dimensional gaze vector Vgaze=(x, y, z), calculated using the following formula:
\begin{equation}
\label{eq3}
\begin{split}
  &x=\ -\cos{\left(pitch\right)}\bullet sin(yaw) \\
 &y=\ -sin(pitch) \\
 &z=-\cos{\left(pitch\right)}\bullet cos(yaw)
\end{split}
\end{equation}

\section{Experimental}
To verify the effectiveness of the EM-Net model proposed in this paper, a large number of experiments are conducted based on publicly available and widely used datasets, mainly including comparative experiments between the EM-Net model and other advanced models. Gaussian noise is added to the dataset to test the robustness of the model, and GAM is compared with SE and CBAM in terms of performance. GAM is visualized to effectively expand the receptive field of the model, and ablation experiments are conducted to explore the impact of key components on the performance of EM-Net. In addition, this section provides a detailed introduction to the experimental details and dataset.
\subsection{Implementation Details }
The experiments in this section are implemented based on the PyTorch framework using the Adam \cite{29} optimizer and the resolution size of the input image is 224 × 224 × 3.The EM-Net model is pre-trained using the dataset ETH-XGaze\cite{30} and the weight parameters obtained are initialized on the dataset MPIIFaceGaze \cite{20}, Gaze360 \cite{31}, and RT-Gene \cite{32} . The initial learning rate is set to 0.0005 and decayed to 0.00025 after 60 epochs.In order to validate the performance of the model, angular error, parameters, FLOPs, and inference time are used as evaluation metrics. Among them, the smaller angular error indicates the better model performance, and the shorter inference time indicates the higher real-time performance of the model. Assuming that the actual gaze direction is $g\in\mathbb{R}^3$, and the predicted gaze direction is $\hat{g}\in\mathbb{R}^3$, the angular error (◦) is calculated as follows:
\begin{equation}
L_{angular}=\arccos(\frac{g\bullet\hat{g}}{\parallel g \parallel\parallel\hat{g}\parallel\ \ })
\end{equation}

MAE Loss (Mean Absolute Error) is used for the loss function, and the formula is as follows:
\begin{equation}
MAE=\ \frac{1}{N}\sum_{i=1}^{N}\left|g_i-{\hat{g}}_i\right|
\end{equation}
Where N represents the total number of samples in the data set.
\subsection{Datasets}
The model performance is verified based on public datasets MPIIFaceGaze \cite{20}, Gaze360 \cite{31}, and RT-Gene \cite{32}. The Gaze360 dataset contains images of 238 subjects in multiple scenarios. The MPIIFaceGaze dataset contains nearly 45k images of 15 subjects which are from the daily life environment, and the samples have the characteristics of long recording time, complex and changeable external environment, diverse eye appearance, and rich lighting conditions, etc. The RT-Gene dataset contains 92K images of 15 subjects. The difference from other datasets used in previous research is that only 50\% of the training data of each dataset is used in this paper, and the interval sampling method is adopted on the original dataset to retain the samples with odd serial numbers so that the sample size of the dataset becomes 50\% of the original.

\begin{table*}[!htbp]
\caption{Comparison with State-of-the-art methods.}\label{tbl2}
\resizebox{1.0\linewidth}{!}{
\begin{tabular}{lcccccc}
\hline
Methods               & MPIIFaceGaze   & Gaze360         & RT-Gene        & Data volume   & Year \\ \hline
Dilated-Net\cite{36}   & 4.42°          & 13.73°          & 8.38°          & 100\%         & 2018 \\
CA-Net\cite{37}       & 4.27°          & 11.20°          & 8.27°          & 100\%         & 2020 \\
SwAT\cite{38}         & 5.00°          & 11.60°          & N/A°           & 100\%         & 2022 \\
GazeTR-Hybrid\cite{35} & 4.00°          & 10.62°          & 6.55           & 100\%         & 2022 \\
CADSE\cite{39}        & 4.04°          & 10.70°          & 7.00°          & 100\%         & 2022 \\
L2CS-Net\cite{21}     & 3.92°          & 10.41°          & N/A            & 100\%         & 2023 \\
GazeCaps\cite{40}      & 4.06°          & \textbf{10.04°} & 6.92°          & 100\%         & 2023 \\
SPMCCA-Net\cite{41}    & N/A            & 10.33°          & 6.68°          & 100\%         & 2023 \\
GazeNAS-ETH\cite{33}   & 3.96°          & 10.52°          & 6.40°          & 100\%         & 2023 \\
Gaze-Swin\cite{34}     & N/A            & 10.14°          & 6.38°          & 100\%         & 2024 \\ \hline
\textbf{EM-Net(Our)}  & \textbf{3.88°} & 10.29°          & \textbf{6.27°} & \textbf{50\%} &      \\ \hline
\end{tabular}
}
\end{table*}

\begin{table}[htbp] 
\caption{Comparison with the state-of-the-art gaze estimation methods in Parameters, FLOPs, and Inference Time.}
\label{tbl3}
\centering
\resizebox{\linewidth}{!}{ 
\begin{tabular}{lccc} 
\toprule
Methods               & Params(M) & FLOPs(G) & Time(ms) \\ 
\midrule
Gaze360\cite{31}       & 11.9      & 7.29     & 49       \\
Dilated-Net\cite{36}   & 3.92      & 3.15     & 23       \\
L2CS-Net\cite{21}      & 23.53     & 4.13     & 27       \\
Gaze-Swin\cite{34}     & 32.28     & 5.17     & N/A      \\
GazeNAS-ETH\cite{33}   & 1.03      & 0.28     & N/A      \\
CADSE\cite{39}         & 74.8      & 19.75    & 67       \\
GazeTR-Hybrid\cite{35} & 11.42     & 1.83     & 19       \\ 
\midrule
EM-Net (Ours)          & \textbf{2.93} & \textbf{0.31} & \textbf{18} \\ 
\bottomrule
\end{tabular}
}
\end{table}

\subsection{Comparison with other methods}
To verify the performance of EM-Net, EM-Net is compared with the existing advanced model, including Angle error, dataset sample size, inference time, Parameters, and FLOPs. The results are shown in \textcolor{blue}{Table \ref{tbl2}} and \textcolor{blue}{Table \ref{tbl3}}. \textcolor{blue}{Table \ref{tbl2}} shows the angle errors of different models on different datasets and the amount of data used to train the models, with EM-Net using only 50\% of the training data. The angle errors of the models included in \textcolor{blue}{Table \ref{tbl2}} all refer to the data in the original paper. \textcolor{blue}{Table \ref{tbl3}} shows the inference time, parameters, and FLOPs of the model. Since the inference time of the model will change with the change of hardware devices, in order to compare the inference time of the model fairly, the data is calculated by running the model under the same hardware conditions. Parameters and FLOPs will not change with the change of different hardware environments, so the data in the original paper is cited.

From \textcolor{blue}{Table \ref{tbl2} and \ref{tbl3}}, it can be seen that although GazeNAS-ETH  \cite{33} has fewer parameters than EM-Net, EM-Net has reduced angle errors by 0.08° (2.02\%), 0.24° (2.2\%), and 0.13° (2.03\%) on the MPIIFaceGaze, Gaze360, and RT-Gene datasets, respectively, and used 50\% less training data than GazeNAS-ETH; Compared with Gaze-Swin  \cite{34}, Gaze-Swin has a 0.15° smaller angle error than EM-Net on the Gaze360 dataset. However, the parameters and FLOPs are 11.02 times and 16.68 times higher than EM-Net. Which is because Gaze-Swin adopts a dual branch structure, consisting of ResNet18 and Swin Transformer, respectively. Compared with GazeTR-Hybrid \cite{35}, EM-Net achievess performance improvement on all datasets with a 74.3\% reduction in parameters and an 83.1\% reduction in FLOPs,  especially on RT-Gene data set with a significant performance improvement of 4.2\%. Although the parameters and FLOPs of EM-Net are significantly reduced, the model inference time is only improved by 1ms. This is because EM-Net integrates the EM algorithm, which requires repeated iterations to converge the model. The entire iteration process is time-consuming, so the overall inference time of the model is not significantly improved. Based on the experimental results and combined with the analysis of the structural characteristics of the EM-Net model, the reasons for the performance improvement of EM Net using only 50\% of training data are as follows:
\begin{itemize} 
\item The GAM proposed in this paper integrates spatial information and channel information, and significantly expands the receptive field of the model, enabling it to capture long-distance dependencies. The GAM can learn the importance weight of each input position, thereby focusing more on the most valuable part of the current task, improving the performance and generalization ability of the model.

\begin{figure*}[htbp]
	\centering
	\includegraphics[width=.8\textwidth]{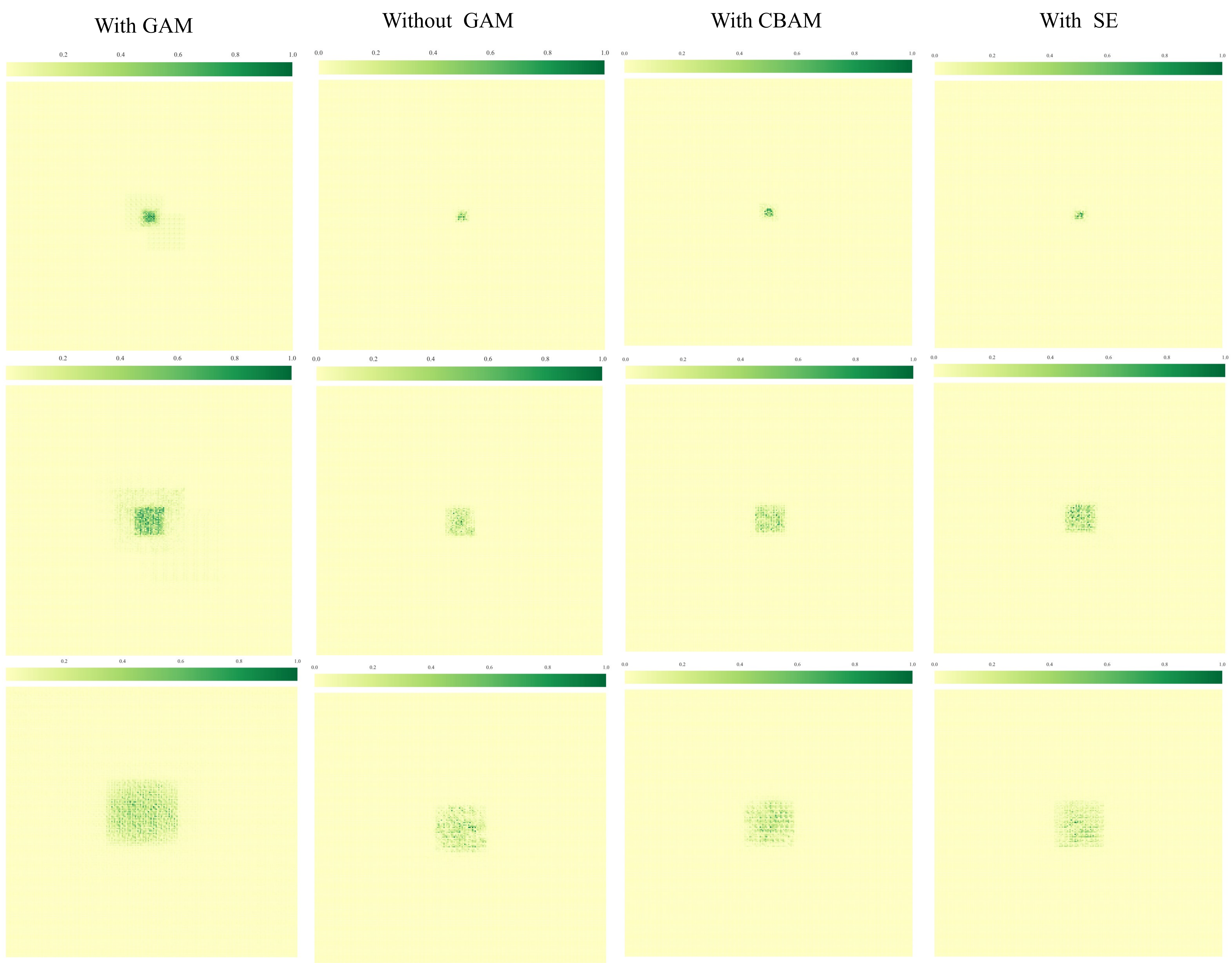}
	\caption{Visualization of receptive field.}
	\label{Receptive_field}
\end{figure*}

\item The EM module proposed in this paper retains the characteristics of the EM algorithm. The EM algorithm can effectively handle the unobserved parts of the data, provide maximum likelihood estimation of parameters, and increase the likelihood function value of the data every iteration, ensuring the convergence of the model. When dealing with incomplete data, the model parameters are updated by calculating the expected value of missing data and using the expected value in reverse.
\item MobileNetV3 itself also has powerful feature extraction capabilities. 
\end{itemize}

\begin{table}[htbp] 
\caption{Validation of GAM effectiveness.}
\label{tbl4}
\centering 
\begin{tabular}{lccc} 
\toprule
Attention         & MPIIFaceGaze   & Gaze360         & RT-Gene \\
\midrule
SE\cite{20}       & 4.02°          & 10.59°          & 6.58°   \\
CBAM\cite{43}     & 3.95°          & 10.62°          & 6.63°   \\
\textbf{GAM(Our)} & \textbf{3.88°} & \textbf{10.29°} & \textbf{6.27°} \\ 
\bottomrule
\end{tabular}
\end{table}

\subsection{Validation of GAM effectiveness}
This section quantitatively evaluates the GAM in EM-Net by replacing with SE and CBAM, while keeping the dataset and hyperparameters unchanged. The detailed data is shown in \textcolor{blue}{Table \ref{tbl4}}. In addition, to observe the impact of different attention mechanisms on the receptive field more intuitively, this section use the receptive field visualization method proposed by Ding et al. \cite{42} to visualize the receptive field of the EM-Net network model after adding different attention mechanisms. The visualization results are shown in \textcolor{blue}{Fig.\ref{Receptive_field}}. The first, second, third, and fourth rows represent the receptive field size of the model without GAM, with GAM, replaced with SE for GAM, and CBAM for GAM, respectively. Each column represents the receptive field at different depths of the model, and detailed data is shown in \textcolor{blue}{Table \ref{tbl1}}.

From \textcolor{blue}{Table \ref{tbl4}} and \textcolor{blue}{Fig. \ref{Receptive_field}}, it can be seen that GAM not only improves model performance but also effectively expands the receptive field of the model. By analyzing the internal structure of GAM, it can be concluded that GAM not only integrates spatial and channel information but also utilizes the self-attention mechanism of shifted windows to fully utilize global contextual information for long-distance modeling, thereby improving overall model performance and expanding receptive fields. CBAM \cite{43} combines channel information and spatial information to improve model performance, but the small receptive field may not fully utilize contextual information, so the performance improvement may not be significant for tasks that require combining contextual information.

\begin{figure*}[htbp]
	\centering
	\includegraphics[width=.9\textwidth]{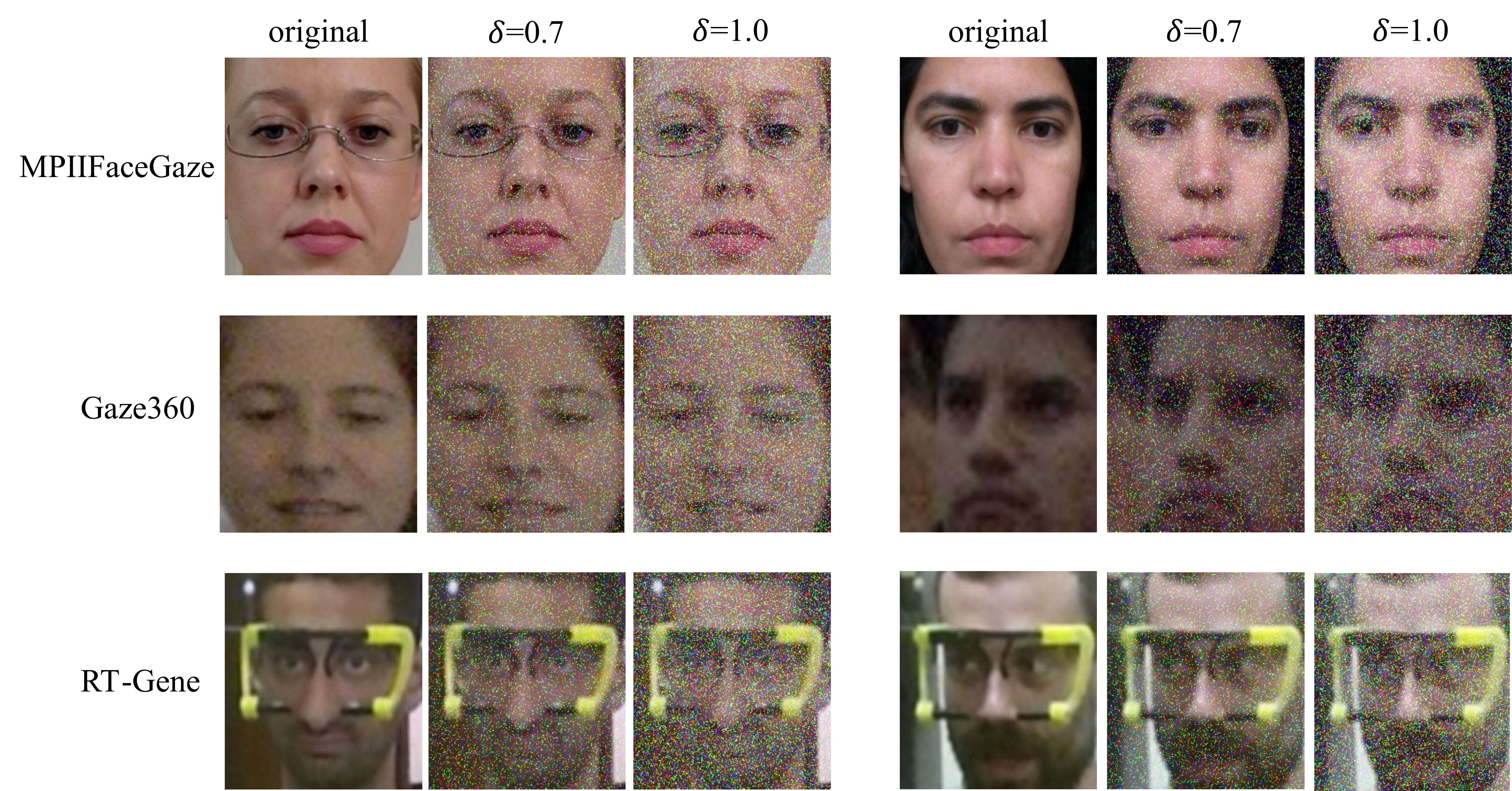}
	\caption{Example of data with Gaussian noise.}
	\label{Gaussian_noise}
\end{figure*}

\begin{table}[h] 
\caption{Robustness testing.}\label{tbl5}
\begin{tabular*}{0.9\linewidth}{@{\extracolsep{\fill}}cccc@{}}
\toprule
Methods   & MPIIFaceGaze & Gaze360 & RT-Gene \\
\midrule
original    & \textbf{3.88°}     & \textbf{10.29°}  & \textbf{6.27°}   \\
$\delta$=0.7 & 4.10°     & 10.69°  & 6.68°   \\
$\delta$=1.0 & 4.45°     & 11.23°  & 6.95°   \\ 
\bottomrule
\end{tabular*}
\end{table}

\subsection{Robustness testing}
The images captured in the real world are not always ideal, and it is easy to produce noise when the image sensor works for a long time, resulting in too high temperature or uneven light during the shooting process. In order to verify whether the performance of the model can maintain robustness in the face of noise, this paper simulates the noise of the actual image by adding Gaussian noise to the image. The probability density of Gaussian noise follows a Gaussian distribution, with two parameters: the mean and the standard deviation. The formula is as follows:
\begin{equation}
\begin{aligned}
f\left(x\right)=\frac{1}{\delta\sqrt{2\pi}}e^{-\frac{{(x-\mu)}^2}{{2\delta}^2}}\\
\delta=\ \sqrt{\frac{\sum_{i=1}^{N}{(X_i-\mu)}^2}{N}}
\end{aligned}
\end{equation}
Among them, $\mu$ represents the mean, and $\delta$ represents the standard deviation.

In order to verify the robustness of the EM-Net model to noise, Gaussian noise with standard deviations of 0.7 and 1.0 is added to the images of the training dataset, respectively, and the data after the addition of Gaussian noise is shown in \textcolor{blue}{Fig. \ref{Gaussian_noise}}. It can be seen from  \textcolor{blue}{Fig. \ref{Gaussian_noise}} that Gaussian noise has a greater impact on data when illumination conditions are poor. To quantitatively analyze the influence of Gaussian noise with different standard deviations on model performance, this paper tests EM-Net Angle errors on MPIIFaceGaze, Gaze360, and RT-Gene datasets after adding Gaussian noise. Detailed data are shown in  \textcolor{blue}{Table \ref{tbl5}}. When the data is added with Gaussian noise, EM-Net can still maintain good performance although there are different degrees of performance degradation. According to the experimental results and theoretical analysis, the reasons why EM-Net can maintain good performance include the following:
\begin{itemize} 
\item CNN can automatically filter out noise and redundant information in the data during feature extraction, thereby improving data quality.
\item The EM module proposed in this paper is able to take into account the effect of noise during the iterative computation, and hence to compute the model parameters more robustly. Moreover, the EM module performs well in dealing with missing data, which can be caused by factors such as noise and occlusion. For example, in the RT-Gene dataset in  \textcolor{blue}{Fig. \ref{Gaussian_noise}}, the eye area is obstructed by glasses.
\item The EM module can effectively model the noise in the graph and reduce the impact of noise on the final result by optimizing model parameters.
\end{itemize}

\subsection{Ablation experiment}
\begin{figure}[htbp]
	\centering
	\includegraphics[width=.35\textwidth]{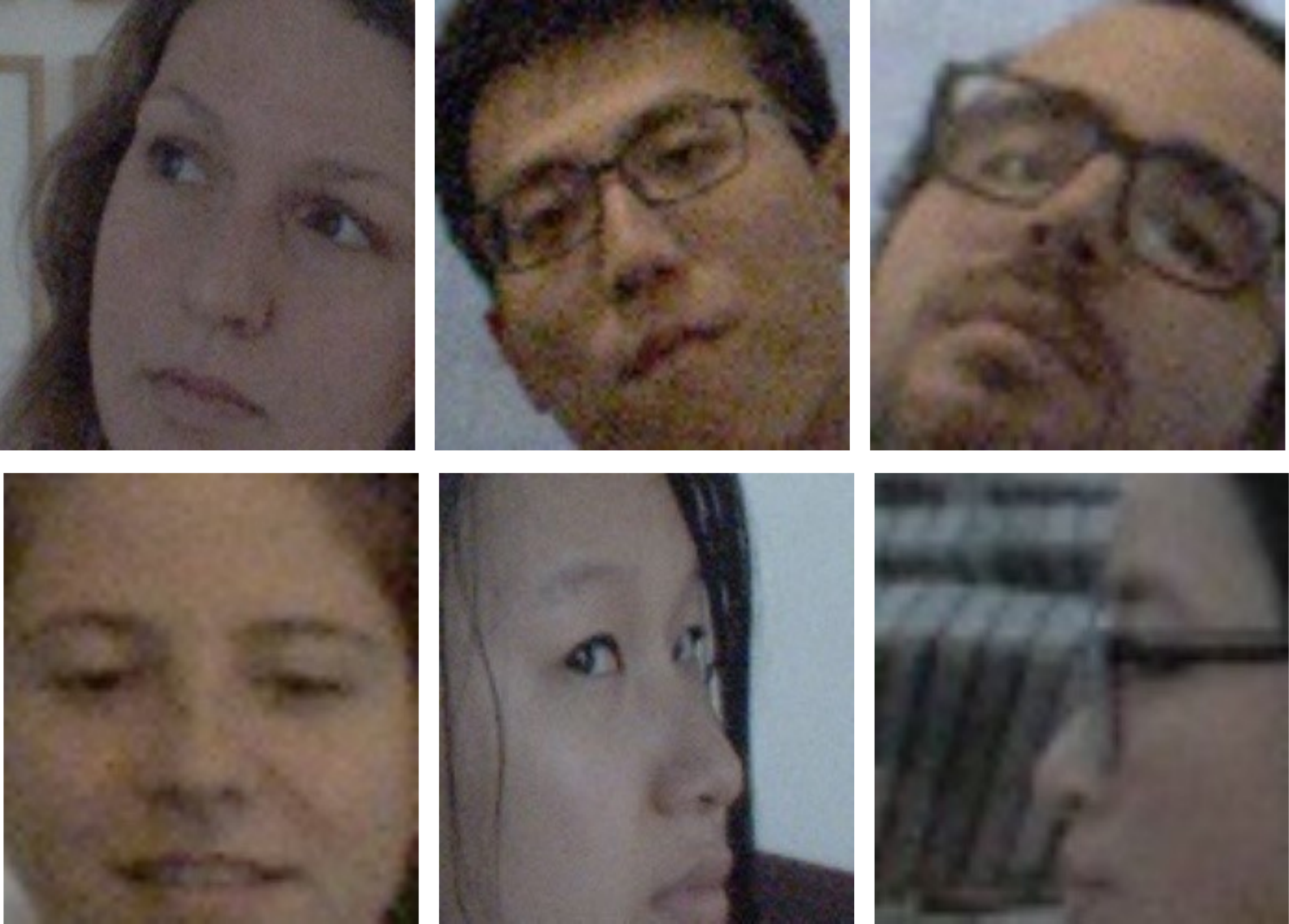}
	\caption{Example of the gaze360 dataset.}
	\label{gaze360}
\end{figure}

In order to better understand the impact of GAM and EM modules on the EM-Net model, ablation experiments are conducted, and the results are shown in \textcolor{blue}{Table \ref{tbl6}}.
It can be seen from  \textcolor{blue}{Table \ref{tbl6}}. that GAM and EM modules have roughly the same impact on model performance. When GAM and EM modules are removed at the same time, the performance of the network model decreases significantly, and the Angle error of the Gaze360 dataset increases by 0.74°(7.2\%). The samples in the Gaze360 dataset have problems such as low clarity and diversified head postures. The diversity of head postures leads to a wider range of head tilt angles and gaze angles.  \textcolor{blue}{Fig. \ref{gaze360}}. shows some samples.

\begin{table}[htbp] 
\caption{Ablation experiment.}
\label{tbl6}
\centering 
\scalebox{0.9}{
\begin{tabular}{ccccc} 
\toprule
GAM	& EM Module   & MPIIFaceGaze & Gaze360 & RT-Gene \\
\midrule
\checkmark  & \checkmark    & \textbf{3.88°}     & \textbf{10.29°}  & \textbf{6.27°}   \\
\checkmark  & $\times$      & 4.02°              & 10.64°           & 6.39°            \\
$\times$    & \checkmark    & 4.00°              & 10.72°           & 6.36°            \\
$\times$    & $\times$      & 4.17°              & 11.03°           & 6.91°            \\
\bottomrule
\end{tabular}}
\end{table}

\subsection{Data visualization}
\begin{figure}[htbp]
	\centering
	\includegraphics[width=.5\textwidth]{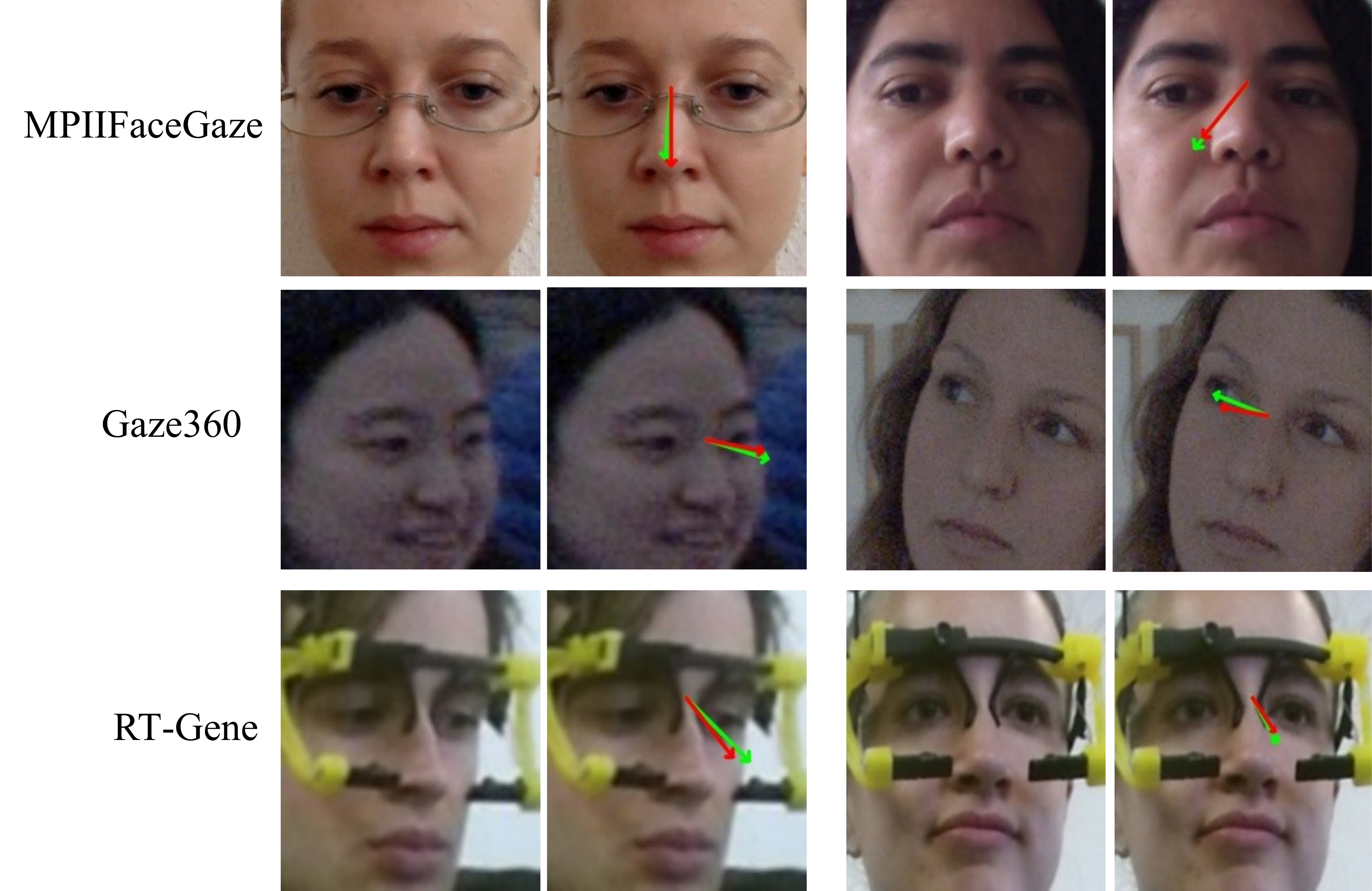}
	\caption{Visualization of gaze estimation direction.}
	\label{Data_visualization}
\end{figure}

In order to provide a more intuitive representation of the gaze estimation results of the EM-Net model on three datasets, \textcolor{blue}{Fig. \ref{Data_visualization}} shows the visualization results of some samples in these three datasets. The green arrow represents the actual gaze direction, the red arrow represents the predicted gaze direction of the model in this paper, and from top to bottom, they represent the MPIIFaceGaze dataset, Gaze360 dataset, and RT-Gene dataset, respectively. The images in \textcolor{blue}{Fig. \ref{Data_visualization}} contain various states such as gender, image quality, head pose, and shot angle. When the eye area of the input facial image is clearly visible and unobstructed (such as the image in the first row), the direction of the model's output gaze is basically consistent with the actual direction. When the image clarity is not high enough, the head rotation angle is large, or the shooting angle is different (such as the image in the second row), there is a certain error between the calculated gaze direction by the model and the actual implementation direction. When the image quality is good and the head rotation angle is small (such as the image in the third row), the angle error of the model is relatively small. This demonstrates that the quality of the input image, changes in lighting, head pose, and shot angle can affect the performance of the model.

\section{Conclusion}
In this paper, we propose a novel lightweight model EM-Net for gaze estimation tasks and validate its performance on three widely used datasets. The experimental results show that EM-Net achieves excellent performance while maintaining lightweight while using only 50\% of the data. EM-Net leverages the Global Attention Mechanism proposed in this paper to integrate channel and spatial information, effectively expanding the receptive field of the model to enable long-distance modeling capabilities. The EM module iteratively computes the model parameters to make the model more robust overall, which improves the overall performance.


\begin{thebibliography}{00}
\bibitem{1}
J.~Li, Z.~Chen, Y.~Zhong, H.-K. Lam, J.~Han, G.~Ouyang, X.~Li, and H.~Liu, ``Appearance-based gaze estimation for asd diagnosis,'' \emph{IEEE Transactions on Cybernetics}, vol.~52, no.~7, pp. 6504--6517, 2022.

\bibitem{2}
C.~S. Indi, V.~Pritham, V.~Acharya, and K.~Prakasha, ``Detection of malpractice in e-exams by head pose and gaze estimation,'' \emph{International Journal of Emerging Technologies in Learning (Online)}, vol.~16, no.~8, p.~47, 2021.

\bibitem{3}
S.~AlKheder, ``Experimental road safety study of the actual driver reaction to the street ads using eye tracking, multiple linear regression and decision trees methods,'' \emph{Expert Systems with Applications}, vol. 252, p. 124222, 2024.

\bibitem{4}
W.~Khan, L.~Topham, H.~Alsmadi, A.~Al~Kafri, and H.~Kolivand, ``Deep face profiler (defap): Towards explicit, non-restrained, non-invasive, facial and gaze comprehension,'' \emph{Expert Systems with Applications}, vol. 254, p. 124425, 2024.

\bibitem{5}
S.~Maiti and A.~Gupta, ``Local eye-net: An attention based deep learning architecture for localization of eyes,'' \emph{Expert Systems with Applications}, vol. 239, p. 122416, 2024.

\bibitem{6}
L.~Jianfeng and L.~Shigang, ``Eye-model-based gaze estimation by rgb-d camera,'' in \emph{Proceedings of the IEEE Conference on Computer Vision and Pattern Recognition Workshops}, 2014, pp. 592--596.

\bibitem{7}
X.~Zhou, H.~Cai, Z.~Shao, H.~Yu, and H.~Liu, ``3d eye model-based gaze estimation from a depth sensor,'' in \emph{2016 IEEE international conference on robotics and biomimetics (ROBIO)}.\hskip 1em plus 0.5em minus 0.4em\relax IEEE, 2016, pp. 369--374.

\bibitem{8}
X.~Zhou, H.~Cai, Y.~Li, and H.~Liu, ``Two-eye model-based gaze estimation from a kinect sensor,'' in \emph{2017 IEEE International Conference on Robotics and Automation (ICRA)}.\hskip 1em plus 0.5em minus 0.4em\relax IEEE, 2017, pp. 1646--1653.

\bibitem{9}
R.~Jafari and D.~Ziou, ``Eye-gaze estimation under various head positions and iris states,'' \emph{Expert Systems with Applications}, vol.~42, no.~1, pp. 510--518, 2015.

\bibitem{10}
X.~Zhang, Y.~Sugano, M.~Fritz, and A.~Bulling, ``Appearance-based gaze estimation in the wild,'' in \emph{Proceedings of the IEEE conference on computer vision and pattern recognition}, 2015, pp. 4511--4520.

\bibitem{11}
------, ``Mpiigaze: Real-world dataset and deep appearance-based gaze estimation,'' \emph{IEEE transactions on pattern analysis and machine intelligence}, vol.~41, no.~1, pp. 162--175, 2017.

\bibitem{12}
Y.~Cheng, F.~Lu, and X.~Zhang, ``Appearance-based gaze estimation via evaluation-guided asymmetric regression,'' in \emph{Proceedings of the European conference on computer vision (ECCV)}, 2018, pp. 100--115.

\bibitem{13}
Y.~Cheng, Y.~Bao, and F.~Lu, ``Puregaze: Purifying gaze feature for generalizable gaze estimation,'' in \emph{Proceedings of the AAAI Conference on Artificial Intelligence}, vol.~36, no.~1, 2022, pp. 436--443.

\bibitem{14}
H.~Balim, S.~Park, X.~Wang, X.~Zhang, and O.~Hilliges, ``Efe: End-to-end frame-to-gaze estimation,'' in \emph{Proceedings of the IEEE/CVF Conference on Computer Vision and Pattern Recognition}, 2023, pp. 2687--2696.

\bibitem{15}
Y.~Hisadome, T.~Wu, J.~Qin, and Y.~Sugano, ``Rotation-constrained cross-view feature fusion for multi-view appearance-based gaze estimation,'' in \emph{Proceedings of the IEEE/CVF Winter Conference on Applications of Computer Vision}, 2024, pp. 5985--5994.

\bibitem{16}
X.~Wu, L.~Li, H.~Zhu, G.~Zhou, L.~Li, F.~Su, S.~He, Y.~Wang, and X.~Long, ``Eg-net: Appearance-based eye gaze estimation using an efficient gaze network with attention mechanism,'' \emph{Expert Systems with Applications}, p. 122363, 2023.

\bibitem{17}
A.~Howard, M.~Sandler, G.~Chu, L.-C. Chen, B.~Chen, M.~Tan, W.~Wang, Y.~Zhu, R.~Pang, V.~Vasudevan \emph{et~al.}, ``Searching for mobilenetv3,'' in \emph{Proceedings of the IEEE/CVF international conference on computer vision}, 2019, pp. 1314--1324.

\bibitem{18}
S.~N. Wadekar and A.~Chaurasia, ``Mobilevitv3: Mobile-friendly vision transformer with simple and effective fusion of local, global and input features,'' \emph{arXiv preprint arXiv:2209.15159}, 2022.

\bibitem{19}
T.~Xu, B.~Wu, R.~Fan, Y.~Zhou, and D.~Huang, ``Fr-net: A light-weight fft residual net for gaze estimation,'' \emph{arXiv preprint arXiv:2305.11875}, 2023.

\bibitem{20}
X.~Zhang, Y.~Sugano, M.~Fritz, and A.~Bulling, ``It's written all over your face: Full-face appearance-based gaze estimation,'' in \emph{Proceedings of the IEEE conference on computer vision and pattern recognition workshops}, 2017, pp. 51--60.

\bibitem{21}
A.~A. Abdelrahman, T.~Hempel, A.~Khalifa, A.~Al-Hamadi, and L.~Dinges, ``L2cs-net: Fine-grained gaze estimation in unconstrained environments,'' in \emph{2023 8th International Conference on Frontiers of Signal Processing (ICFSP)}.\hskip 1em plus 0.5em minus 0.4em\relax IEEE, 2023, pp. 98--102.

\bibitem{22}
X.~Cai, J.~Zeng, S.~Shan, and X.~Chen, ``Source-free adaptive gaze estimation by uncertainty reduction,'' in \emph{Proceedings of the IEEE/CVF Conference on Computer Vision and Pattern Recognition}, 2023, pp. 22\,035--22\,045.

\bibitem{23}
A.~Vaswani, N.~Shazeer, N.~Parmar, J.~Uszkoreit, L.~Jones, A.~N. Gomez, {\L}.~Kaiser, and I.~Polosukhin, ``Attention is all you need,'' \emph{Advances in neural information processing systems}, vol.~30, 2017.

\bibitem{24}
J.~Hu, L.~Shen, and G.~Sun, ``Squeeze-and-excitation networks,'' in \emph{Proceedings of the IEEE conference on computer vision and pattern recognition}, 2018, pp. 7132--7141.

\bibitem{25}
Z.~Yang, L.~Zhu, Y.~Wu, and Y.~Yang, ``Gated channel transformation for visual recognition,'' in \emph{Proceedings of the IEEE/CVF conference on computer vision and pattern recognition}, 2020, pp. 11\,794--11\,803.

\bibitem{26}
G.~E. Hinton, S.~Sabour, and N.~Frosst, ``Matrix capsules with em routing,'' in \emph{International conference on learning representations}, 2018.

\bibitem{27}
X.~Zhang, J.~Liu, S.~Chang, P.~Gong, Z.~Wu, and B.~Han, ``Mirn: A multi-interest retrieval network with sequence-to-interest em routing,'' \emph{Plos one}, vol.~18, no.~2, p. e0281275, 2023.

\bibitem{28}
Z.~Liu, Y.~Lin, Y.~Cao, H.~Hu, Y.~Wei, Z.~Zhang, S.~Lin, and B.~Guo, ``Swin transformer: Hierarchical vision transformer using shifted windows,'' in \emph{Proceedings of the IEEE/CVF international conference on computer vision}, 2021, pp. 10\,012--10\,022.

\bibitem{29}
D.~P. Kingma and J.~Ba, ``Adam: A method for stochastic optimization,'' \emph{arXiv preprint arXiv:1412.6980}, 2014.

\bibitem{30}
X.~Zhang, S.~Park, T.~Beeler, D.~Bradley, S.~Tang, and O.~Hilliges, ``Eth-xgaze: A large scale dataset for gaze estimation under extreme head pose and gaze variation,'' in \emph{Computer Vision--ECCV 2020: 16th European Conference, Glasgow, UK, August 23--28, 2020, Proceedings, Part V 16}.\hskip 1em plus 0.5em minus 0.4em\relax Springer, 2020, pp. 365--381.

\bibitem{31}
P.~Kellnhofer, A.~Recasens, S.~Stent, W.~Matusik, and A.~Torralba, ``Gaze360: Physically unconstrained gaze estimation in the wild,'' in \emph{Proceedings of the IEEE/CVF international conference on computer vision}, 2019, pp. 6912--6921.

\bibitem{32}
T.~Fischer, H.~J. Chang, and Y.~Demiris, ``Rt-gene: Real-time eye gaze estimation in natural environments,'' in \emph{Proceedings of the European conference on computer vision (ECCV)}, 2018, pp. 334--352.

\bibitem{33}
V.~Nagpure and K.~Okuma, ``Searching efficient neural architecture with multi-resolution fusion transformer for appearance-based gaze estimation,'' in \emph{Proceedings of the IEEE/CVF winter conference on applications of computer vision}, 2023, pp. 890--899.

\bibitem{34}
R.~Zhao, Y.~Wang, S.~Luo, S.~Shou, and P.~Tang, ``Gaze-swin: Enhancing gaze estimation with a hybrid cnn-transformer network and dropkey mechanism,'' \emph{Electronics}, vol.~13, no.~2, p. 328, 2024.

\bibitem{35}
Y.~Cheng and F.~Lu, ``Gaze estimation using transformer,'' in \emph{2022 26th International Conference on Pattern Recognition (ICPR)}.\hskip 1em plus 0.5em minus 0.4em\relax IEEE, 2022, pp. 3341--3347.

\bibitem{36}
Z.~Chen and B.~E. Shi, ``Appearance-based gaze estimation using dilated-convolutions,'' in \emph{Asian Conference on Computer Vision}.\hskip 1em plus 0.5em minus 0.4em\relax Springer, 2018, pp. 309--324.

\bibitem{37}
Y.~Cheng, S.~Huang, F.~Wang, C.~Qian, and F.~Lu, ``A coarse-to-fine adaptive network for appearance-based gaze estimation,'' in \emph{Proceedings of the AAAI Conference on Artificial Intelligence}, vol.~34, no.~07, 2020, pp. 10\,623--10\,630.

\bibitem{38}
A.~Farkhondeh, C.~Palmero, S.~Scardapane, and S.~Escalera, ``Towards self-supervised gaze estimation,'' \emph{arXiv preprint arXiv:2203.10974}, 2022.

\bibitem{39}
J.~O~Oh, H.~J. Chang, and S.-I. Choi, ``Self-attention with convolution and deconvolution for efficient eye gaze estimation from a full face image,'' in \emph{Proceedings of the IEEE/CVF Conference on Computer Vision and Pattern Recognition}, 2022, pp. 4992--5000.

\bibitem{40}
H.~Wang, J.~O. Oh, H.~J. Chang, J.~H. Na, M.~Tae, Z.~Zhang, and S.-I. Choi, ``Gazecaps: Gaze estimation with self-attention-routed capsules,'' in \emph{Proceedings of the IEEE/CVF conference on computer vision and pattern recognition}, 2023, pp. 2668--2676.

\bibitem{41}
C.~Yan, W.~Pan, C.~Xu, S.~Dai, and X.~Li, ``Gaze estimation via strip pooling and multi-criss-cross attention networks,'' \emph{Applied Sciences}, vol.~13, no.~10, p. 5901, 2023.

\bibitem{42}
X.~Ding, X.~Zhang, J.~Han, and G.~Ding, ``Scaling up your kernels to 31x31: Revisiting large kernel design in cnns,'' in \emph{Proceedings of the IEEE/CVF conference on computer vision and pattern recognition}, 2022, pp. 11\,963--11\,975.

\bibitem{43}
S.~Woo, J.~Park, J.-Y. Lee, and I.~S. Kweon, ``Cbam: Convolutional block attention module,'' in \emph{Proceedings of the European conference on computer vision (ECCV)}, 2018, pp. 3--19.

\end{thebibliography}

\end{document}